\documentclass[runningheads]{llncs}

\usepackage[T1]{fontenc}
\usepackage{boldline}

\usepackage[binary-units=true, detect-all]{siunitx}[=v2]

\usepackage{amsfonts}
\usepackage{amssymb}
\usepackage{amsmath}
\usepackage{graphicx}
\usepackage{xcolor}
\usepackage{soul}
\usepackage{cite}

\begin{document}

\title{Vision-Based Neurosurgical Guidance: Unsupervised Localization and Camera-Pose Prediction \thanks{This work funded by the SNSF (Project IZKSZ3\_218786).}}

\titlerunning{Vision-Based Neurosurgical Guidance}

\author{Gary Sarwin\inst{1} \and
Alessandro Carretta\inst{2,3} \and
Victor Staartjes\inst{2}\and
Matteo Zoli\inst{3}\and
Diego Mazzatenta\inst{3}\and
Luca Regli\inst{2}\and
Carlo Serra\inst{2} \and 
Ender Konukoglu\inst{1}}

\authorrunning{G Sarwin et al.}
\institute{Computer Vision Lab, ETH Zurich, Switzerland \and
Department of Neurosurgery, University Hospital of Zurich, Zurich, Switzerland \and 
Department of Biomedical and Neuromotor Sciences (DIBINEM), University of Bologna, Bologna, Italy}

\maketitle
\begin{abstract}
Localizing oneself during endoscopic procedures can be problematic due to the lack of distinguishable textures and landmarks, as well as difficulties due to the endoscopic device such as a limited field of view and challenging lighting conditions. 
Expert knowledge shaped by years of experience is required for localization within the human body during endoscopic procedures. 
In this work, we present a deep learning method based on anatomy recognition, that constructs a surgical path in an unsupervised manner from surgical videos, modelling relative location and variations due to different viewing angles. 
At inference time, the model can map an unseen video's frames on the path and estimate the viewing angle, aiming to provide guidance, for instance, to reach a particular destination. 
We test the method on a dataset consisting of surgical videos of transsphenoidal adenomectomies, as well as on a synthetic dataset. An online tool that lets researchers upload their surgical videos to obtain anatomy detections and the weights of the trained YOLOv7 model are available at: \url{https://surgicalvision.bmic.ethz.ch}.

\keywords{Neuronavigation  \and Unsupervised Embedding \and Anatomical Recognition \and Endoscopic Surgeries \and Surgical AI \and Surgical Vision}
\end{abstract}
\section{Introduction}
The environment during an endoscopic procedure poses numerous challenges to the surgeons. 
Successfully navigating this environment requires extensive experience combined with an extremely high level of anatomical understanding from the video feed. 
Challenges stem from the inherent nature of the human anatomy, such as non-rigid deformations, the absence of obvious boundaries between anatomical structures, and adverse events like bleeding that can happen during surgery, as well as the imaging device, such as limited field of view and light reflection. 

A variety of methods have been developed to assist neurosurgeons orient themselves during neurosurgeries. 
While computer-assisted neuronavigation has been a crucial tool and long-term research focus in the community \cite{Hartl2013WorldwideSurgery,  Orringer2012NeuronavigationTrends
}, it relies on preoperative imaging and brain shift hampers this reliability~\cite{Iversen2018AutomaticNeuronavigation}. 

Additional real-time anatomical guidance can be achieved through interoperative MRI \cite{Berkmann2014IntraoperativeAdenoma, Staartjes2021MachineSurgery}, ultrasound \cite{Ulrich2012ResectionUltrasound, Burkhardt2014High-frequencyApproach}, the use of fluorescent substances \cite{Stummer2017RandomizedGliomas, Hadjipanayis2015WhatGliomas}, awake surgery \cite{Hervey-Jumper2015AwakePeriod}, and electrophysiological neuromonitoring \cite{DeWittHamer2012ImpactMeta-analysis,Sanai2008FunctionalResection}.
These techniques are efficient, relying on physical traits, however, also costly as they demand proficiency in a new imaging modality and, more importantly, requiring temporary surgery halts or instrument retractions for intraoperative information \cite{Staartjes2020MachineSurvey}.

The pursuit of a more cost-effective real-time solution, independent of additional machinery, coupled with advancements in deep learning techniques, has propelled the development of vision-based localization methods. 
Various approaches, including structure from motion and simultaneous localization and mapping (SLAM) \cite{Grasa2011EKFSequences, Ozyoruk2021EndoSLAMVideos, Mahmoud2016ORBSLAM-basedReconstruction, Leonard2018EvaluationData}, aim for 3D map reconstruction based on feature correspondence. 
\cite{Ozyoruk2021EndoSLAMVideos}.
Many vision-based localization methods rely on distinctive landmark positions and tracking them across frames for localization. 
Factors inherent to endoscopic neurosurgical videos, such as low texture, a lack of distinguishable features, non-rigid deformations, and disruptions \cite{Ozyoruk2021EndoSLAMVideos}, degrade their performance. 
Consequently, alternative solutions are imperative to tackle these challenges.

Despite recent progress, the task of detecting or segmenting anatomical structures, which could serve as a foundation for an alternative approach to neuronavigation, remains under-explored and poses an ongoing challenge.
It is important to note that recognizing anatomy in surgical videos is more challenging than detecting surgical tools, given the absence of clear boundaries and variations in color or texture between anatomical structures. 

Interest in applying machine learning to neurosurgery has increased, especially in pituitary surgery, and first works have explored the possibility of detection and segmentation of anatomical structures \cite{Sarwin2023Unsupervised, Das2023Multitask}. 

\cite{Sarwin2023Unsupervised} reported promising results of anatomical structure detection, and additionally, demonstrated a way to use the detections and their constellation to construct a common surgical path in an unsupervised manner, allowing relative localization during a surgery. 
Furthermore, in \cite{Das2023Multitask} the authors proposed a multi-task network to identify critical structures during the sellar phase of pituitary surgery. Their model, PAINet, jointly predicts segmentation of the two largest structures and centroids of the smaller and less frequently occurring structures. 

In this work, we extend the model proposed in \cite{Sarwin2023Unsupervised} to include the \emph{viewing direction of the endoscope} in the construction of the surgical path and mapping on the path. 
The viewing direction plays an important role in navigation during surgery since it heavily influences the structures and their constellations viewed by the endoscope. 
Thus, including it in the model has the potential to yield better surgical paths and more accurate guidance. 
Importantly, we include the viewing direction, represented as rotational angles around the $x$ and $y$ axis (i.e., pitch and yaw), in the model also in an unsupervised way without assuming the presence of any camera parameters for training.

The unsupervised learning of the surgical path and viewing direction is facilitated by an Autoencoder (AE) architecture using a training set of videos. 
The AE is trained to reconstruct bounding boxes of a given frame based on a constrained latent representation. 
At inference time, the latent representation of a new frame provides relative positioning and viewing angles. 
Unlike approaches reconstructing a 3D environment and relying on landmarks for localization, our method aims to construct a common surgical roadmap and localizing within that map relying on bounding box detections.
Relying on semantic bounding box detections, eliminates the need for tracking arbitrary landmarks, facilitating handling disruptions during surgery, such as bleeding, flushing and retractions. 
The learned mapping relies on the principle that the visibility, relative sizes and constellations of anatomical structures, which can be inferred from bounding box detections, strongly correlate with the position along the surgical trajectory and the viewing angle of the endoscope. 

The proposed approach is demonstrated on the transsphenoidal adenomectomy procedure, chosen for its relatively one-dimensional surgical path. This choice makes it well-suited for proving the concept of our suggested method.

\section{Methods}
\subsection{Problem Formulation and Approach}
\label{sec:ProblemForm}
Let $\mathbf{S}_{t}$ denote a sequence of endoscopic video frames $\mathbf{x}_{t-s:t}$, as illustrated in Figure~\ref{fig:fig2}. 
Here, $s$ denotes the sequence length, and $\mathbf{x}_t \in \mathbb{R}^{w \times h \times c}$ represents the $t$-th frame with $w$, $h$, and $c$ indicating the width, height, and number of channels, respectively. 
Our primary goal is to embed the sequence $\mathbf{S}_{t}$ into a 3D latent dimension represented by the variable $\mathbf{z}=\left[z^1, z^2, z^3\right]$. 

Our approach involves identifying anatomical structures in the sequence $\mathbf{S}_t$ and mapping the frame $\mathbf{x}_t$ to the latent space using the identified structures. 
Notably, $z^1$ serves as an \emph{implicit} surgical atlas, signifying a path from the beginning of the procedure until the final desired anatomy. 
It is implicit because position information along the surgical path is unavailable for constructing the latent space. 
$z^2$ and $z^3$ represent pitch and yaw angles, respectively, forming a rotation matrix to predict the endoscope's viewing direction. Note that depth and camera pose information is rarely available in standard configurations, either because it is inaccessible or the functionality is missing. Extracting and collecting this data from medical devices can be cumbersome, or not possible due to the highly regulated environment and restrictions concerning modifications to these devices. Therefore, these are modeled and learned in an unsupervised way.

\begin{figure}[!b]
    \centering
    \includegraphics[width=0.95\textwidth]{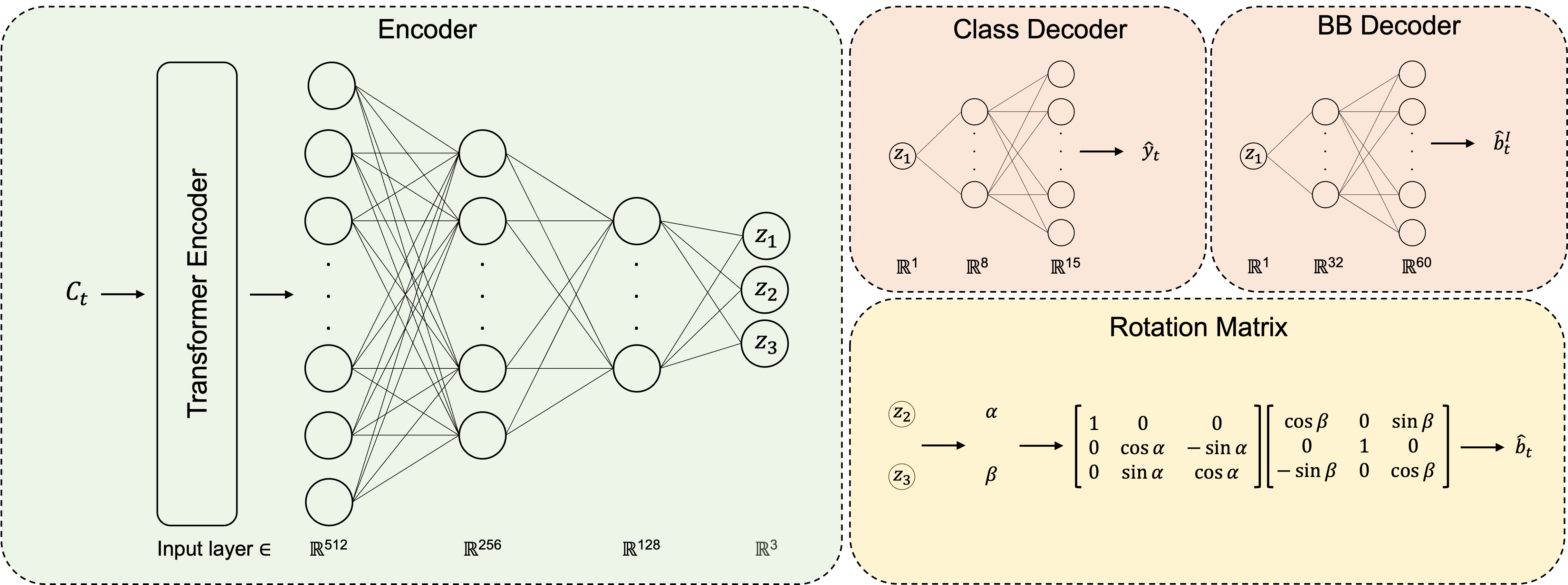}
    \caption{The model comprises an encoder and a decoder that consists of two fully connected networks. The encoder takes $\mathbf{C}_t$ as input and embeds this sequence into a 3D latent representation. The decoder consists of two fully connected networks to generate the class probabilities $\hat{\mathbf{y}}_t$ and the bounding box coordinates $\hat{\mathbf{b}}^I_t$ from $z_t^1$. Furthermore, the encoder outputs $z_t^2$ and $z_t^3$ that are used to construct a rotation matrix to rotate the predicted bounding boxes around the pitch and yaw axes.}
    \label{fig:model}
\end{figure}

To achieve this, object detection is performed on all frames $\mathbf{x}_{t-s:t}$ in $\mathbf{S}_{t}$, resulting in a sequence of detections $\mathbf{c}_{t-s:t}$ denoted as $\mathbf{C}_t$. A detection $\mathbf{c}_t \in \mathbb{R}^{n \times 5}$ includes binary variables $\mathbf{y}_t = [y_t^0,\dots,y_t^n] \in {0,1}^{n}$ indicating presence of structures in the $t$-th frame and bounding box coordinates $\mathbf{b}_t = [\mathbf{b}_t^0,\dots,\mathbf{b}_t^n]^T \in \mathbb{R}^{n \times 4}$. 
An autoencoder architecture is employed to embed $\mathbf{C}_t$ into $\mathbf{z}_t=\left[z_t^1, z_t^2, z_t^3\right]$. The encoder maps $\mathbf{C}_t$ to $\mathbf{z}_{t}$, and the decoder reconstructs $\hat{\mathbf{c}}_t = \left(\hat{y}_t, \hat{b}_t\right)$, representing detections for the last frame in a given sequence. 
$z^1_t$ is used to generate $\hat{y}_t$ and $\hat{b}^I_t=\left[\hat{b}_t^{I,0},\dots,\hat{b}_t^{I,n}\right]^T$, the bounding box location reconstruction assuming pitch and yaw angles are zero, i.e., in a \emph{centered view}.
$z^2_t$ and $z^3_t$ are used to build a rotation matrix $\mathbf{R}_t$, as shown in Figure \ref{fig:model}, to model variability due to differences in viewing angles. The final bounding box reconstructions, $\hat{b}_t$, are obtained by rotating the bounding-box center coordinates of $\hat{b}^I_t$ with $\mathbf{R}_t$ and keeping the same bounding-box size in the rotated position. 

The model parameters are updated to ensure that the prediction with the rotated center coordinates, $\hat{\mathbf{c}}_t$, fits $\mathbf{c}_{t}$ in a training set, as elaborated in the following.

\subsection{Object Detection}
Our method involves the identification of anatomical structures by detecting them as bounding boxes in video frames. For this purpose, we utilize the YOLOv7 network in the object detection phase of our pipeline \cite{Wang2022YOLOv7:Detectors}. The network is trained on endoscopic videos from a training set, where frames are sparsely labeled with bounding boxes. 
Subsequently, the trained network is applied to all frames of the training videos to generate detections for these classes on every frame. These detections serve as the input for training the subsequent autoencoder, which models the embedding. For further processing, the instrument class was omitted as this class does not necessarily correlate with the position on the surgical path.

\subsection{Embedding and Camera-Pose}
The encoder of the AE comprises multi-head attention layers followed by fully connected layers, ultimately reducing the input to 3 latent dimensions, where $z_t^1$ represents the position on the surgical path, and $z_t^2$ and $z_t^3$ represent pitch and yaw angles, which represent the rotation angles of the camera with respect to a centered view. A transformer-based encoder is employed to encode the temporal information in the sequence of detections. 

The decoder consists of two fully connected networks, the first generating the class probabilities $\hat{\mathbf{y}}_t$ of $\hat{\mathbf{c}}_t$, and the second generating the corresponding bounding boxes $\hat{\mathbf{b}}^I_t$ from $z_t^1$ in the centered view.
To obtain bounding box reconstructions in the observed view, the center coordinates of the predicted bounding boxes $\hat{\mathbf{b}}^I_t$ are rotated by multiplying them with the rotation matrix $\hat{\mathbf{R}}_t$, as explained in Section \ref{sec:ProblemForm}, to obtain $\hat{\mathbf{b}}_t$, the second component of $\hat{\mathbf{c}}_t$. 
The AE is designed to reconstruct only the last frame $\mathbf{c}_{t}$ in $\mathbf{C}_t$ since $z_{t}^1$ is intended to correspond to the current position. 
However, it considers $s$ previous frames to provide additional information in determining the latent representation $\mathbf{z}_t$ of $\mathbf{x}_t$.

The loss function consists of a classification loss and a bounding box loss, the latter being calculated only for the classes present in the ground truth. In the current setup, the fully connected network that produces the bounding boxes can classify any view as the centered view, since the bounding box coordinates can be rotated to fit the input even if the centered view is not at pitch and yaw angles of zero. Therefore, for increased interpretability, we enforce that the output is centered by feeding the output of the bounding box network together with $\mathbf{y}_t$ once again into the encoder by stacking the output $s$-times to achieve the same input size, and add the predicted $\mathbf{\hat{z}}_{t}^2$ and $\mathbf{\hat{z}}_{t}^3$, which should both be zero if the view is centered, to the loss function. 
This leads to the objective to minimize for the $t$-th frame in the $m$-th training video:
\begin{equation*}
\begin{aligned}
\mathcal{L}_{m, t}= & -\sum_{i=1}^n\left(y_{m, t}^i \log \left(\hat{y}_{m, t}^i\right)+\left(1-y_{m, t}^i\right) \log \left(1-\hat{y}_{m, t}^i\right)\right) \\
& +\sum_{i=1}^n y_{m, t}^i\left|\mathbf{b}_{m, t}^i-\hat{\mathbf{b}}_{m, t}^{i}\right|+\sqrt{\left(\hat{z}_{m, t}^2\right)^2}+\sqrt{\left(\hat{z}_{m, t}^3\right)^2},
\end{aligned}
\end{equation*}
where $|\cdot|$ denotes the $l_1$ loss.

The total training loss is then the sum of $\mathcal{L}_{m,t}$ over all frames and training videos. The proposed loss function can be viewed as maximizing the joint likelihood of a given $\mathbf{y}$ and $\mathbf{b}$ with a probabilistic model utilizing a mixture model for the bounding boxes. 

\section{Experiments and Results}
\subsection{Datasets}
\begin{figure}[!t]
\centering
\includegraphics[width=0.95\textwidth]{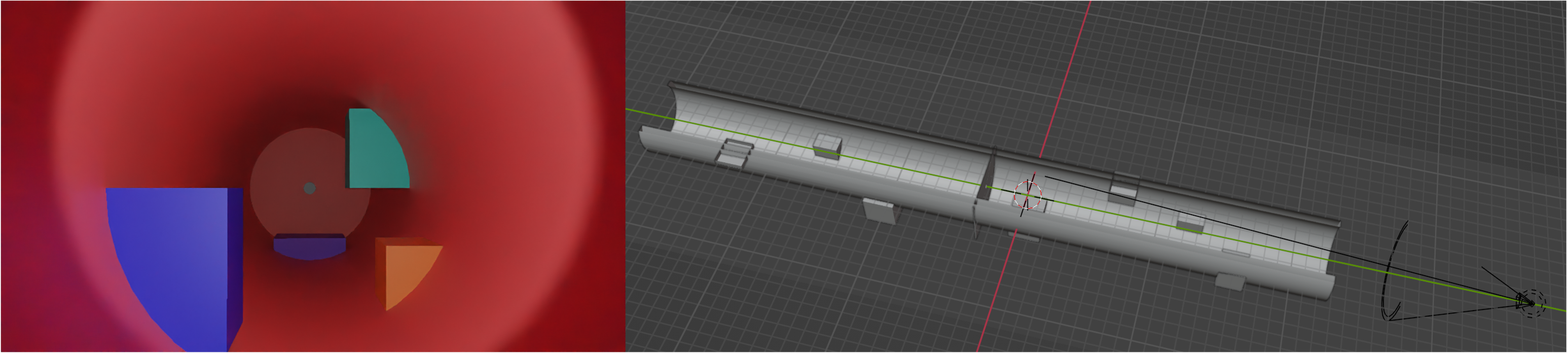}
\caption{An overview of the model used for the creation of the synthetic dataset.}
\label{fig:fig2}
\end{figure}

In this work two datasets were utilized, a medical dataset and a synthetic dataset. \\
\emph{\bf Medical Dataset:}
The dataset utilized for object detection contains 166 videos documenting transsphenoidal adenomectomy procedures in 166 patients. These videos, captured using a variety of endoscopes across multiple facilities over 10 years, were made accessible under general research consent. Expert neurosurgeons annotated the videos, encompassing 16 distinct classes, namely, 15 anatomical structure classes and one class for surgical instruments. The dataset encompasses approximately 19,000 labeled frames. Each class has one instance per video, except for the instrument class since various instruments are categorized under the same class. Among the 166 videos, 146 were allocated for training and validation purposes, while the remaining 20 were purposed for testing. Despite the utilization of data from various centers, it is important to acknowledge potential biases introduced by the geographical vicinity of these centers. 
\emph{\bf Synthetic Dataset:}
For quantitative analysis, a synthetic dataset was created in Blender \cite{Blender} with ground truth. A 3D environment was built to represent a surgical path with various structures. Ground-truth object detection labels could be extracted from the software. Eight different objects were modeled, akin to different anatomical structures, with a single instance per object to emulate the medical setting. To train the AE, a video moving through the environment was recorded with random viewing directions, moving forward and backward several times. In total, the data consists of 16502 frames with corresponding ground-truth object detection labels. The model is depicted in Figure \ref{fig:fig2}.

\subsection{Implementation Details}

The YOLO network was trained with identical parameters and implementation as in \cite{Sarwin2023Unsupervised}, which follows standard implementation as reported in \cite{Wang2022YOLOv7:Detectors}.

The AE integrates a transformer encoder comprising six transformer encoder layers, each with five heads, and an input size of $s \times 15 \times 5$, where $s$ is established as 64 frames. Following this, the output dimension of the transformer encoder undergoes reduction through three fully connected layers to sizes of 512, 256, and 128, respectively, employing rectified linear unit (ReLU) activation functions between layers. Subsequently, the final fully connected layer reduces the dimensionality to $\mathbf{z}_t \in \mathbb{R}^{3}$, employing a sigmoid activation function to obtain $z_t^1$, and a tanh activation function for $z_t^2$ and $z_t^3$. Moreover, the decoder composed of two fully connected networks, namely the class decoder and bounding box decoder, have two fully connected layers, elevating the dimensionality of the latent variable $z_t^1$ from 1D to 8, to $\mathbf{n}$, and from 1D to 32, to $\mathbf{n} \times 4$, correspondingly. The initial layer of both of those networks is succeeded by a ReLU activation function, while the final layer adopts a sigmoid activation function. Furthermore,  for ${z}_t^2$ and ${z}_t^3$, an output of $-1$ and $1$ represents a $-90$ and $90$ degrees rotation around the respective axis, these outputs are then converted to radians for the construction of the rotation matrix. 

For the AE's training, the AdamW optimizer \cite{Loshchilov2017DecoupledRegularization} is employed in conjunction with a warm-up scheduler, which linearly increases the learning rate from 0 to \num{1e-4} over a span of 60 epochs. The model is trained for a total of 2500 epochs for the synthetic, and 270 epochs for the medical dataset. The model has approximately 4.6M parameters in the setup for the medical dataset. 
\subsection{Results}
{\bf Anatomical Detection:}
The YOLOv7 trained on the medical dataset reaches a mean average precision 53.4\% at a 0.5 intersection over union threshold as also was reported in~\cite{Sarwin2023Unsupervised}. \\
{\bf Quantitative Assessment of the Embedding:}
Due to the absence of all camera parameters, the exact modeling of rotation is an ill-posed task. However, we show that even though we introduce significant simplifications and substitute the homography matrix with a straightforward rotation matrix, we can still approximate the angles of rotation. We test on 1022 sequences which were recorded in Blender under random viewing angles moving through the model. 
We report a mean error in angle predictions of 0.43, and 0.69 with a standard deviation of 2.38 and 1.74 degrees, for the pitch and yaw angle, respectively.

Additionally, we examine the correlation between the predicted location along the surgical path with the real depth of the synthetic model for a video traveling through the model. We do this for both our AE and the model proposed in \cite{Sarwin2023Unsupervised}. 
We expect our AE that takes rotation into account to embed space more representative compared to an AE that does not consider rotation. 
More specifically, our AE should be able to map depth to the surgical path, and account for different views by taking the rotation into account, whereas the previous model that only maps to the surgical path without rotation would need to occupy more space in the 1D latent space to describe various views of the same location and thereby describing depth less accurately. 
Our AE achieved a Pearson correlation coefficient of 0.97 whereas the previous model proposed in \cite{Sarwin2023Unsupervised} achieved 0.94.\\
{\bf Qualitative Assessment of the Embedding:}
\begin{figure}[t]
\centering
\includegraphics[width=0.95\textwidth]{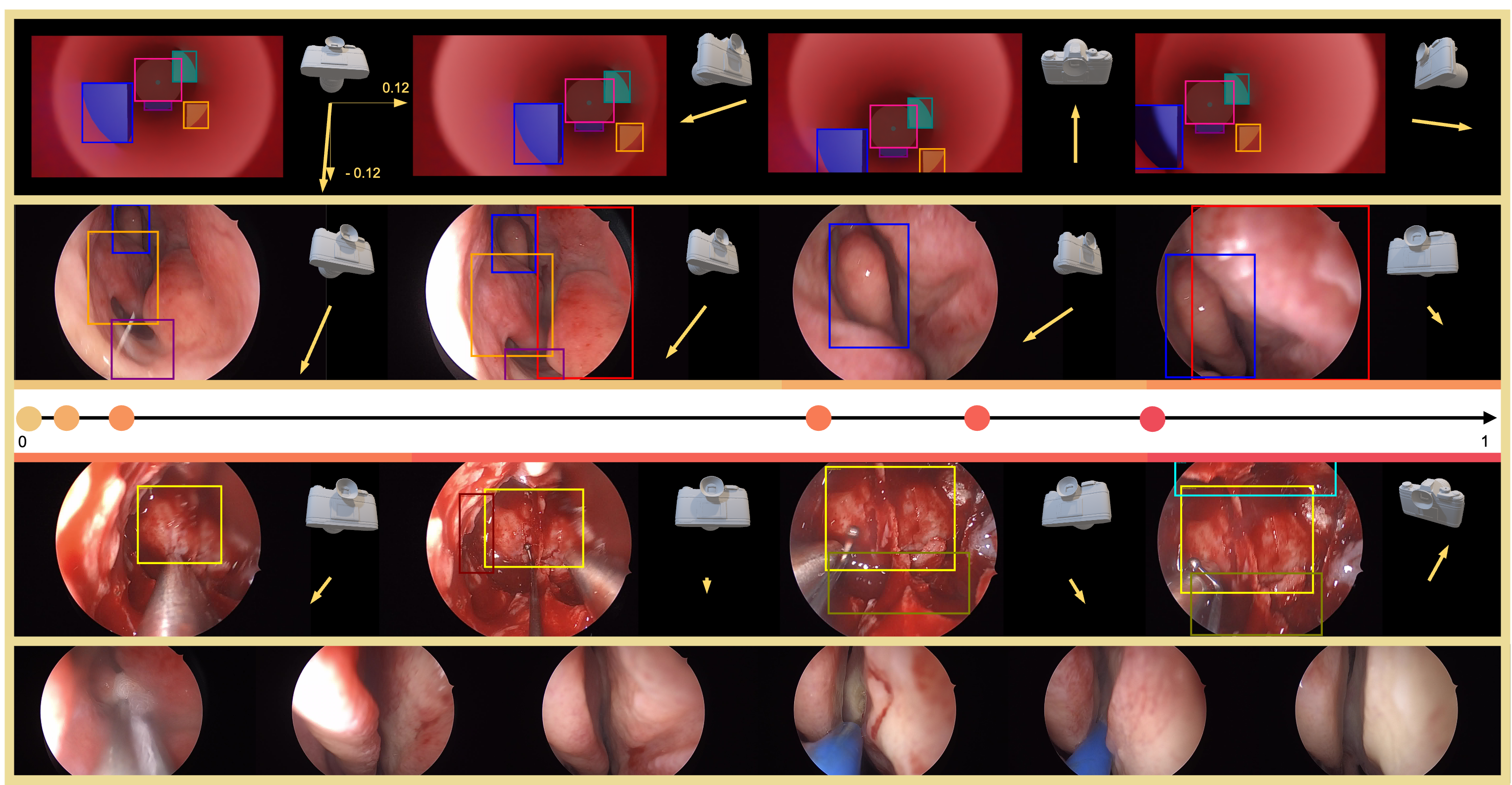}
\caption{Results are shown that depict the predicted viewing direction of the model for the sequences in the synthetic dataset (row 1), as well as the medical dataset (row 2,3). Additionally, for the medical dataset, their predicted location along the surgical path is shown. The depicted cameras are for illustrative purposes only. Finally, in the bottom row images are shown that are mapped to the same location along the surgical path by the AE. We can see the same anatomical location under different point of views and during different stages of the surgery. }
\label{fig:fig3}
\end{figure}
Of greater importance is that the model can tell the surgeon whether the viewing direction is right, or whether the camera should be pointed in another direction, and additionally can tell the surgeon whether in that direction should be looked more or less. In Figure \ref{fig:fig3} sequences are shown for the medical dataset, together with the predictions $\mathbf{z}_t$. The arrows plot the negative predictions of ${z}_t^2$ and ${z}_t^3$ to visualize the general direction the camera is pointed in, and the magnitude. For illustrative purposes a camera model is depicted to visualize the camera's orientation. The sequences visualize the ability of the model to extract various viewing directions from the images which correspond to the movement of the camera between the images in one sequence and that the predicted viewing directions are in line with the movement of visual landmarks between images within a sequence. The idea is that by having a reference visualization of the desired or planned viewing direction that is required to find a certain anatomical structure or for orientation purposes, i.e. an arrow as in Figure \ref{fig:fig3}, the surgeons can adjust the current orientation of the endoscope with respect to the reference arrow, without the need for exact angles, but rather which direction it should be pointed in, and more or less so. Videos visualizing the results are supplied in the supplementary material.

Finally, in the bottom row of Figure \ref{fig:fig3}, we depict images that are mapped to the same location by our AE. These images show the same anatomical position during different stages of the surgery as well as from different viewpoints. When encoding the same images using the AE proposed in \cite{Sarwin2023Unsupervised}, these images are mapped within a range of 0.81\% of the latent space instead of a single point. This demonstrates that the proposed AE that takes rotation into account can embed space more coherently and confirms the results of quantitative correlation experience.

\section{Conclusion}

In this study, we introduced an approach to neuronavigation leveraging deep learning techniques. 
Our proposed method is image-based and utilizes bounding box detections of anatomical structures to orient itself within a surgical path learned from a dataset of surgical videos, and additionally provides feedback on the direction the camera, i.e. endoscope, is pointing in. 
This is facilitated through an AE architecture trained without supervision. 
Our approach enables the localization and prediction of anatomical structures along the surgical trajectory, both forward and backward, similar to functionalities seen in mapping applications. 

However, our work also comes with certain limitations. Primarily, we have confined our focus to a single surgical procedure in this preliminary investigation. 
Extending to other surgical procedures is a goal for future research. 
Additionally, our proposed method can be integrated with SLAM techniques, as well as leveraging guidance from pre- or intra-operative MRI. 
Both of these extensions constitute areas of future exploration. 
Another limitation lies in the fact that the latent dimension only offers relative positional and angular encoding. 
To surpass this limitation, additional labeling of the actual positions along the surgical path may be necessary, as well as camera parameters.

\bibliographystyle{splncs04}

\end{document}